\documentclass[10pt,twocolumn,letterpaper]{article}
\usepackage{cvpr}
\usepackage{times}
\usepackage{epsfig}
\usepackage{graphicx}
\usepackage{amsmath}
\usepackage{amssymb}
\usepackage{hyperref}
\usepackage{booktabs,multirow, multicol}
\usepackage{graphicx}
\usepackage{subcaption}

\usepackage{xcolor}
\usepackage{colortbl}

\definecolor{Gray}{rgb}{0.9, 0.9, 0.9}
\usepackage{pifont}

\begin{document}

\title{\LARGE \bf xEdgeFace: Efficient Cross-Spectral Face Recognition for Edge Devices}
\author{Anjith George and S\'ebastien Marcel \\
Idiap Research Institute \\
Rue Marconi 19, CH - 1920, Martigny, Switzerland \\
{\tt\small  \{anjith.george, sebastien.marcel\}@idiap.ch  }
}

\maketitle
\thispagestyle{empty}

\begin{abstract}

Heterogeneous Face Recognition (HFR) addresses the challenge of matching face images across different sensing modalities, such as thermal to visible or near-infrared to visible, expanding the applicability of face recognition systems in real-world, unconstrained environments. While recent HFR methods have shown promising results, many rely on computation-intensive architectures, limiting their practicality for deployment on resource-constrained edge devices. In this work, we present a lightweight yet effective HFR framework by adapting a hybrid CNN-Transformer architecture originally designed for face recognition. Our approach enables efficient end-to-end training with minimal paired heterogeneous data while preserving strong performance on standard RGB face recognition tasks. This makes it a compelling solution for both homogeneous and heterogeneous scenarios. Extensive experiments across multiple challenging HFR and face recognition benchmarks demonstrate that our method consistently outperforms state-of-the-art approaches while maintaining a low computational overhead. 

\end{abstract}

\section{Introduction}
Facial recognition (FR) has become a key component in modern biometric systems, especially for access control, thanks to its efficiency and non-intrusive nature. With the rise of deep learning, particularly convolutional neural networks (CNNs), face recognition has reached near-human performance under unconstrained conditions~\cite{learned2016labeled}. However, most of these systems are built for homogeneous settings, where both gallery and probe images are captured using visible-spectrum cameras.

In many real-world scenarios, such as surveillance, mobile authentication, or defense applications, relying solely on visible-light imagery is limiting. Images captured beyond the visible spectrum, such as near-infrared (NIR) \cite{happy2012video} or thermal, offer clear advantages. For instance, NIR is more robust to changes in lighting and is harder to spoof~\cite{li2007illumination,george2022comprehensive}. Despite these benefits, training effective models on such modalities remains a challenge due to the scarcity of large-scale, annotated heterogeneous datasets. Heterogeneous Face Recognition (HFR) aims to bridge the gap between different sensing modalities, such as matching a thermal or NIR image to a visible-light reference~\cite{klare2012heterogeneous,anghelone2025beyond}. A key subtask here is Cross-spectral Face Recognition (CFR), which deals with extreme appearance differences caused by spectral shifts between domains. CFR is especially crucial in low-light or long-range environments where visible imaging is impractical. Beyond the spectrum, HFR also covers matching across sketches, depth, 2D-3D data, or even low-resolution inputs.

While recent advances in deep neural networks (DNNs) have significantly improved Heterogeneous Face Recognition (HFR), the task remains challenging due to the inherent modality gap between source and target domains, which often causes RGB-trained models to perform poorly on non-RGB data~\cite{he2018wasserstein}. Moreover, collecting large-scale, paired cross-modal datasets is both costly and logistically demanding, highlighting the need for models that generalize well under limited supervision. At the same time, the heavy architectures used in many state-of-the-art HFR methods hinder deployment on edge devices. This has led to a growing interest in lightweight models that strike a balance between efficiency and accuracy. Recent developments in Vision Transformers (ViTs), known for capturing long-range dependencies~\cite{khan2022transformers}, present a compelling alternative to CNNs. Combining the strengths of both paradigms opens new possibilities for building compact yet robust HFR systems suitable for real-world, resource-constrained environments.

In this work, we introduce a novel HFR framework based on a hybrid CNN-Transformer architecture EdgeFace \cite{george2023edgeface}, originally developed for RGB-based face recognition. Our approach starts from a backbone pre-trained on large-scale RGB datasets and adapts it to cross-modal settings with minimal heterogeneous supervision. Unlike existing methods that require extensive paired data or large models, our solution enables efficient end-to-end training in a lightweight design suited for real-time applications.

Our model delivers strong performance across both homogeneous and heterogeneous settings. It maintains competitive accuracy on standard RGB benchmarks while outperforming existing methods on cross-modal datasets. We validate our approach through comprehensive experiments on popular FR as well as HFR benchmarks. 

\noindent The main contributions of this work are: \begin{itemize} 
\item We propose a framework to train a lightweight hybrid CNN-Transformer architecture for Heterogeneous Face Recognition, enabling robust cross-modal matching with a minimal amount of paired data. 
\item Our model is designed for efficiency, making it practical for deployment on resource-constrained edge devices. 
\item Extensive experiments across multiple challenging HFR and RGB benchmarks show that our approach achieves competitive or superior performance compared to state-of-the-art methods while being extremely lightweight. 
\end{itemize}

\section{Related works}

\textbf{Heterogeneous Face Recognition}: HFR aims to match facial images captured under different sensing modalities, such as visible (VIS), near-infrared (NIR), thermal, or sketch domains. A major challenge in HFR lies in the modality gap, i.e., the large distribution shift between modalities, which significantly degrades recognition performance when conventional face recognition models trained on RGB data are applied directly to new modalities. To mitigate this, recent research has proposed a wide range of techniques broadly categorized into three main paradigms: invariant feature learning, common-space projection, and synthesis-based approaches.

Invariant feature-based methods aim to extract robust facial descriptors that remain consistent across modalities. Early works leveraged handcrafted features such as Difference of Gaussian (DoG) filters, multi-scale LBP \cite{liao2009heterogeneous}, SIFT, and MLBP \cite{klare2010matching} to model local texture patterns. Later approaches incorporated deep CNNs to learn modality-invariant embeddings \cite{he2017learning,he2018wasserstein}, while others introduced novel handcrafted descriptors like the Local Maximum Quotient (LMQ) \cite{roy2018novel} or composite feature integration at the score level \cite{liu2018composite}.

Common-space projection methods attempt to reduce domain discrepancies by mapping multi-modal features into a shared latent space. Classical methods include Canonical Correlation Analysis (CCA) \cite{yi2007face}, Partial Least Squares (PLS) \cite{sharma2011bypassing}, and coupled regression models \cite{lei2009coupled}. These techniques align modalities through linear or nonlinear transformations that preserve discriminative information. More recent methods adopt deep learning-based solutions, such as domain-specific units \cite{de2018heterogeneous}, domain-invariant units \cite{george2024heterogeneous},  coupled attribute-guided loss functions \cite{liu2020coupled}, and semi-supervised collaborative representations \cite{liu2023modality}, which improve cross-domain alignment with minimal manual supervision or paired data. Recent works \cite{george2024modalities, george2023bridging} have shown that conditional modulation of the intermediate feature maps could address the domain gap, which was later extended to be modality agnostic \cite{george2024modality}.

Synthesis-based methods take a different route by generating modality-translated images, often in the visible domain, so that standard face recognition pipelines can be directly applied. Early techniques relied on patch-level synthesis using Markov Random Fields \cite{wang2008face} or manifold learning methods like LLE \cite{liu2005nonlinear}. The introduction of GANs and CycleGAN \cite{zhuUnpairedImagetoImageTranslation2017} has significantly advanced this line of research, enabling unpaired image translation and photo-realistic face synthesis \cite{zhang2017generative,fu2021dvg}. Recent innovations include latent disentanglement models \cite{liu2021heterogeneous}, memory-modulated transformers for unsupervised reference-based generation \cite{luo2022memory}, and plug-and-play modules like Prepended Domain Transformers (PDT) \cite{george2022prepended}, which align cross-domain features without explicit synthesis. However, the synthesize-based methods increase the computation required heavily as we need to use both image translation and another face recognition network for matching. 

\textbf{Lightweight Face Recognition}: With the widespread adoption of handheld mobile devices and edge computing, the focus in face recognition (FR) research has shifted toward developing lightweight models that maintain high accuracy while operating under strict computational constraints. This has led to a surge of interest in efficient network designs tailored for FR. MobileFaceNets~\cite{chen2018mobilefacenets}, based on the MobileNet architecture~\cite{howard2017mobilenets,sandler2018mobilenetv2}, were among the first to demonstrate high accuracy with fewer than 1M parameters. MixFaceNets~\cite{boutros2021mixfacenets} adopted the MixConv concept~\cite{tan2019mixconv} to further improve efficiency, and ShiftFaceNet~\cite{wu2018shift} leveraged ShiftNet’s operations to reach competitive performance with just 0.78M parameters. ShuffleFaceNet~\cite{martindez2019shufflefacenet}, inspired by ShuffleNetV2~\cite{ma2018shufflenet}, provided flexible model sizes from 0.5M to 4.5M parameters while maintaining high accuracy. Architecture search has also played a key role: PocketNet~\cite{boutros2022pocketnet} was derived using DARTS on CASIA-WebFace~\cite{yi2014learning}, employing multi-step knowledge distillation (KD), while VarGFaceNet~\cite{yan2019vargfacenet}, the winner of the ICCV 2019 LFR challenge~\cite{deng2019lightweight}, used variable group convolutions to balance computational load. More recently, SynthDistill~\cite{shahreza2024knowledge} showed that synthetic data \cite{george2025digi2real} and online KD can effectively train TinyFaR networks~\cite{han2020model}, allowing student models to mimic powerful teacher networks. GhostFaceNets~\cite{alansari2023ghostfacenets} exploited redundancy in convolutional operations to build compact models with as few as 61M FLOPs using depthwise convolutions. Finally, EdgeFace \cite{george2024edgeface} fused convolutional and transformer modules inspired by EdgeNeXt~\cite{maaz2023edgenext}, incorporating low-rank linear modules to reduce both parameter count and computational cost, achieving near state-of-the-art performance at a fraction of the complexity, ranking first among the compact models in the EFaR 2023 challenge  \cite{kolf2023efar}. 

\textbf{Lightweight Heterogeneous Face Recognition}: Lightweight face recognition (FR) models are well-suited for edge deployment, yet their extension to the more challenging Heterogeneous Face Recognition (HFR) task remains underexplored. Existing HFR methods often rely on computationally heavy architectures or synthesis-based pipelines that introduce prohibitive computational overhead, limiting real-world applicability on resource-constrained devices. To bridge this gap, we propose a compact and efficient HFR framework designed for edge devices, achieving strong cross-modal performance with minimal computational and data requirements.
 
\section{Proposed Approach}

\begin{figure*}[t!]
  \centering
      \includegraphics[width=0.95\linewidth]{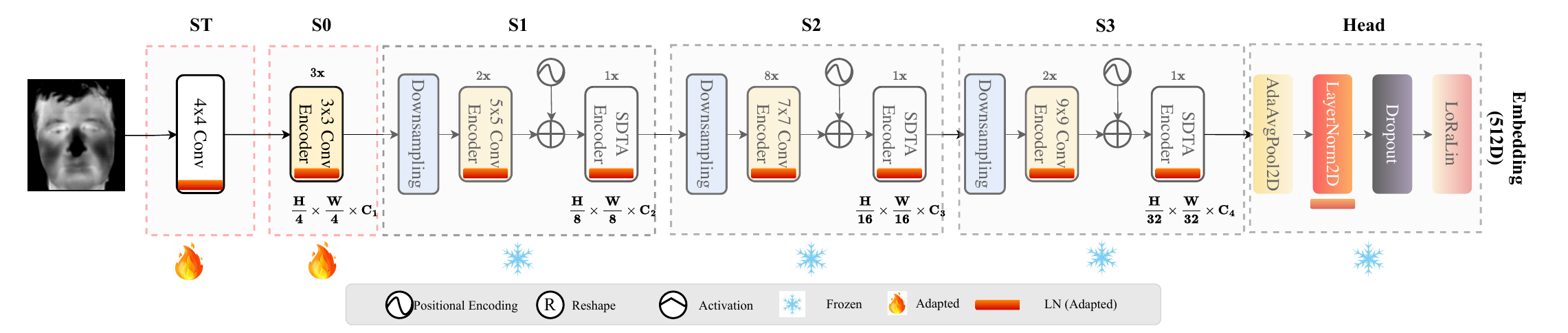}
  \caption{Model architecture of xEdgeFace models: The highlighted modules (LN-LayerNorm, ST-Conv. Stem, Stages-S0, S1, S2) are adapted while other network components remain frozen. The two loss components ensure modality alignment, keeping the source FR performance. Computational complexity remains unchanged in new models.}
  \label{fig:framework}
\end{figure*}

Heterogeneous face recognition (HFR) presents a unique challenge due to the scarcity of paired training data. A commonly adopted strategy to mitigate this issue involves leveraging large-scale pretrained models on visible-spectrum (RGB) data and adapting them to the heterogeneous domain. However, fine-tuning such models directly on the small HFR datasets often leads to overfitting and catastrophic forgetting, where the model's original RGB face recognition capability is significantly degraded.
To address this, several prior works \cite{de2018heterogeneous,george2022prepended} have proposed architectural modifications that introduce modality-specific branches or asymmetric processing paths. Although such designs help preserve RGB performance, they come at the cost of increased model complexity and parameter redundancy, an issue particularly detrimental when targeting lightweight models.
Moreover, these approaches often assume a fixed representation for the RGB modality, requiring the heterogeneous domain (e.g., NIR, sketch, or thermal) to map into the same latent space. This rigid alignment can become a performance bottleneck, especially when modality-specific variations are large and nonlinear.
In this work, we aim to design a unified model capable of handling both homogeneous and heterogeneous face recognition tasks without incurring significant compute overhead or compromising performance in either domain. Specifically, we propose a lightweight yet effective adaptation strategy for pretrained face recognition models that avoids catastrophic forgetting while enabling robust cross-modal generalization.
We base our approach on EdgeFace \cite{george2023edgeface}, a state-of-the-art lightweight architecture that combines convolutional backbones with transformer-based components. A key insight in our method is the pivotal role of Layer Normalization (LayerNorm \cite{gatys2016image}) in modality adaptation. Rather than altering the backbone or duplicating pathways, we utilize LayerNorm as a modulation point to adapt modality-specific statistics, allowing the network to learn discriminative features for both domains within a shared architecture.
To achieve this, we utilize a contrastive self-distillation training framework. The training objective consists of two components: 1) Contrastive Modality Alignment, which enforces that embeddings from paired samples (e.g., RGB-NIR) are close in the feature space, promoting modality-invariant representations. 2) Self-Distillation Regularization, which maintains the original RGB recognition capabilities by distilling knowledge from the pretrained model into the adapted one, thus mitigating catastrophic forgetting.

Together, these components allow us to fine-tune the model on limited HFR data while preserving performance on the original RGB task and maintaining architectural efficiency. Detailed descriptions of the training objectives, backbone,  and implementation details are provided in the following subsections.

\textbf{LayerNorm Adaptation}: Previous studies have demonstrated that the statistical properties of feature maps in deep neural networks (DNNs) can effectively capture the stylistic elements of images, as first illustrated by Gatys et al.~\cite{gatys2016image}. Building on this understanding of internal representations, normalization techniques have become essential for stabilizing and improving the training of deep models. Layer Normalization (LayerNorm), introduced by Ba et al.~\cite{ba2016layer}, addresses the limitations of batch normalization by computing normalization statistics across the features of individual samples rather than across the batch dimension. This design ensures consistent behavior during both training and inference, making it especially suitable for scenarios involving variable input sizes or non-i.i.d. data. All hidden units within a layer share the same normalization parameters, though these parameters vary across samples, enabling more fine-grained control over neuron activations. In the context of large language models (LLMs) and their multi-modal extensions (MLLMs), recent work by Zhao et al.~\cite{zhao2024tuning} has shown that fine-tuning LayerNorm parameters within each attention block can significantly enhance model efficiency while reducing computational overhead, outperforming other parameter-efficient approaches such as Low-Rank Adaptation (LoRA) \cite{hu2022lora}. LayerNorm’s capacity to mitigate internal covariate shift, accommodate diverse data modalities, and facilitate stable optimization makes it particularly well-suited for adapting face recognition (FR) models to heterogeneous face recognition (HFR) scenarios.

\textbf{Problem Formulation}: We begin with a pretrained face recognition network $F$, parameterized by weights $\Theta_{\text{FR}}$, trained on a large-scale dataset in the visible (RGB) spectrum. Let $(X_{s_i}, X_{t_i}, y_i)$ denote a triplet consisting of a pair of images $X_{s_i}$ and $X_{t_i}$ from the source (e.g., RGB) and target (e.g., NIR or thermal) modalities respectively, and a binary identity label $y_i \in \{0, 1\}$, where $y_i = 1$ indicates that both images correspond to the same identity and $y_i = 0$ otherwise.

The goal is to adapt $F$ into a heterogeneous face recognition network $\hat{F}$, parameterized by $\Theta_{\text{HFR}}$, such that the resulting embeddings $e_{s_i} = \hat{F}(X_{s_i})$ and $e_{t_i} = \hat{F}(X_{t_i})$ are well-aligned in the shared embedding space if they belong to the same identity, while also preserving the discriminative ability of the original model $F$ on the source modality.

We initialize $\hat{F}$ with the pretrained parameters $\Theta_{\text{FR}}$, and decompose $\Theta_{\text{HFR}}$ into three disjoint subsets:
\begin{equation}
\Theta_{\text{HFR}} = \left\{ \Theta_{\text{LN}}^{(1:K)}, \Theta_{\text{Adapted}}, \Theta_{\text{Frozen}} \right\},
\end{equation}
where $\Theta_{\text{LN}}^{(1:K)}$ denotes the set of all LayerNorm parameters (from $K$ layers), $\Theta_{\text{Adapted}}$ includes all trainable parameters except LayerNorms, and $\Theta_{\text{Frozen}}$ refers to the set of parameters that remain fixed during training.

To enforce alignment between embeddings from different modalities, we use a cosine-based contrastive loss defined as follows:
\begin{equation}
\label{eq:contrastive}
\begin{aligned}
\mathcal{L}_{\text{C}}(e_{s_i}, e_{t_i}, y_i) =\; & y_i \cdot \left(1 - \cos(e_{s_i}, e_{t_i})\right) \\
& + (1 - y_i) \cdot \max\left(0, \cos(e_{s_i}, e_{t_i}) - m\right),
\end{aligned}
\end{equation}
where $\cos(e_{s_i}, e_{t_i}) = \frac{e_{s_i} \cdot e_{t_i}}{\|e_{s_i}\|_2 \|e_{t_i}\|_2}$ is the cosine similarity between the embeddings and $m \in [0, 1]$ is a contrastive margin.

To prevent the model from forgetting the original face recognition capability on the source modality, we introduce a self-distillation loss that encourages the adapted model $\hat{F}$ to preserve the source modality embeddings of the pretrained model $F$. This is defined as:
\begin{equation}
\label{eq:self_distillation}
\mathcal{L}_{\text{SDL}}(e_{F_{s_i}}, e_{\hat{F}_{s_i}}) = 1 - \cos(e_{F_{s_i}}, e_{\hat{F}_{s_i}}),
\end{equation}
where $e_{F_{s_i}} = F(X_{s_i})$ is the frozen embedding from the original model and $e_{\hat{F}_{s_i}} = \hat{F}(X_{s_i})$ is the adapted embedding for the same image.

The final loss function for optimizing the adapted network $\hat{F}$ is a weighted combination of the contrastive loss for modality alignment and the self-distillation loss for performance preservation:
\begin{equation}
\label{eq:total_loss}
\begin{aligned}
\mathcal{L}_{\text{total}} = &\; (1 - \lambda) \cdot \mathcal{L}_{\text{C}}(e_{s_i}, e_{t_i}, y_i) \\
& + \lambda \cdot \mathcal{L}_{\text{SDL}}(e_{F_{s_i}}, e_{\hat{F}_{s_i}}),
\end{aligned}
\end{equation}
where $\lambda \in [0, 1]$ is a balancing hyperparameter controlling the trade-off between cross-modal alignment and self-regularization.

In all our experiments, we set $\lambda = 0.75$ and margin $m = 0$ unless stated otherwise. This configuration empirically provided the best balance between adapting to the heterogeneous domain and retaining source modality performance.

\textbf{Face Recognition Backbone} We adopt the pre-trained EdgeFace \cite{george2024edgeface} model as the face recognition (FR) backbone. EdgeFace is a hybrid convolutional-transformer architecture that incorporates LayerNorm modules in place of the more commonly used BatchNorm. The model is trained on the WebFace12M dataset \cite{zhu2021webface260m}, comprising over 12 million images from more than 600,000 unique identities. Input images are RGB with a spatial resolution of $112 \times 112$ pixels. Before inference, all face images are aligned using a similarity transformation to normalize eye positions to fixed coordinates. For thermal (single-channel) inputs, the image is triplicated across channels to produce a three-channel tensor suitable for processing by the RGB-trained backbone.

\textbf{Implementation Details}. Our HFR framework leverages a frozen copy of the pre-trained EdgeFace model as a regularizer network to guide the learning of a shallow, trainable surrogate network (Figure \ref{fig:framework}) through self-distillation. Specifically, the surrogate is initialized with the pretrained model weights and fine-tuned on the target domain. During training, only select lower-level modules, including LayerNorm layers, are unfrozen for adaptation, while the rest of the network remains fixed. To enhance cross-modal alignment, we integrate a contrastive loss applied to the feature embeddings produced by the surrogate model across both modalities (RGB and thermal).

The framework is implemented in PyTorch and builds on the Bob library \cite{bob2012, bob2017}\footnote{\url{https://www.idiap.ch/software/bob/}}. We use the Adam optimizer with a learning rate of $1 \times 10^{-4}$, training for 20 epochs with a batch size of 256. The margin parameter $m$ in the contrastive loss is set to 0, and  $\lambda$ is fixed at 0.75 across all experiments. Although both the pretrained and surrogate networks are used during training, only the adapted surrogate needs to be retained for inference.

\section{Experiments}
This section presents the results from a comprehensive set of experiments conducted using the proposed framework. We assessed the heterogeneous face recognition (HFR) performance on standard benchmarks and compared it to state-of-the-art methods in the literature. Additionally, we tested the framework on standard face recognition benchmarks to ensure that adaptation did not result in catastrophic forgetting. For all experiments, the standard cosine distance metric was employed for performance comparison.

\subsection{Datasets and Protocols}
For our evaluations, we utilized the following datasets:

\textbf{Tufts face Dataset:} The Tufts Face Database \cite{panetta2018comprehensive} comprises a diverse collection of face images across various modalities, designed for HFR tasks. In our evaluation, we specifically utilized the thermal images available in the dataset following the VIS-Thermal protocol. This dataset includes 113 identities, with a demographic breakdown of 39 males and 74 females from various regions. Each subject is represented across different modalities. Following the protocol established in \cite{fu2021dvg}, we randomly selected 50 identities for the training set and allocated the remaining subjects to the test set.

\textbf{MCXFace Dataset:} The MCXFace Dataset \cite{george2022prepended,mostaani2020high} comprises images of 51 individuals captured under varying illumination conditions across three distinct sessions and utilizing multiple channels. These channels include RGB color, thermal, near-infrared (850 nm), short-wave infrared (1300 nm), depth, and depth estimated from RGB images. The dataset features five different folds, each created by randomly dividing the subjects into training and development partitions. Our evaluations focused on the challenging ``VIS-Thermal'' protocols of this dataset.

\textbf{Polathermal Dataset:} The Polathermal dataset \cite{hu2016polarimetric}, collected by the U.S. Army Research Laboratory (ARL), is a heterogeneous face recognition (HFR) dataset that includes both polarimetric LWIR imagery and color images for 60 subjects. Additionally, the dataset provides both conventional thermal images and polarimetric images for each subject. In our experiments, we utilized the conventional thermal images, following the five-fold partitioning method introduced in \cite{de2018heterogeneous}. We used 25 identities for the training set and the remaining 35 identities for the test set.

\textbf{SCFace Dataset:} The SCFace dataset \cite{grgic2011scface} features high-quality enrollment images for face recognition alongside low-quality probe samples taken in diverse surveillance settings using various cameras. This dataset is organized into four protocols that vary based on the quality and distance of the probe samples: close, medium, combined, and far, with the ``far'' protocol posing the greatest challenge. Overall, the dataset comprises 4,160 static images from 130 subjects, captured across both visible and infrared spectra.

\textbf{CUFSF Dataset:} The CUHK Face Sketch FERET Database (CUFSF) \cite{zhang2011coupled} is composed of 1,194 face images from the FERET dataset \cite{phillips1998feret}, each paired with a corresponding sketch created by an artist. The artistic exaggerations in the sketches make this dataset particularly challenging for heterogeneous face recognition (HFR) tasks. Following the protocols in \cite{fang2020identity}, we trained our model using 250 identities and tested it on the remaining 944 identities.

\textbf{CASIA NIR-VIS 2.0 Dataset:} The CASIA NIR-VIS 2.0 Face Database \cite{li2013casia} includes images of 725 individuals captured under both visible spectrum and near-infrared (NIR) lighting conditions. Each individual in the dataset is represented by 1-22 visible spectrum photos and 5-50 NIR photos. The experimental setup employs a 10-fold cross-validation approach, with 360 identities designated for training. The gallery and probe sets for evaluation contain 358 unique individuals, ensuring that the identities in the training and testing sets are completely distinct. 

\textbf{Metrics}: Our evaluation of the models employs a range of performance metrics that are well-established in prior literature. These include the Area Under the Curve (AUC), Equal Error Rate (EER), Rank-1 identification rate, and Verification Rate at various false acceptance rates (0.01\%, 0.1\%, 1\%, and 5\%). When multiple folds are present, we report the mean and the standard deviation along the folds.
\subsubsection{\textbf{Ablation Studies}}
Given the large set of design choices and hyperparameters involved, we first conduct a comprehensive set of ablation studies to analyze the contribution of individual components to the overall performance of the model. All ablation experiments are performed on the Tufts Face Dataset using the VIS-Thermal protocol, which presents the most challenging heterogeneous face recognition (HFR) scenario due to its substantial domain gap. Adapted variants of the base models are denoted as \textbf{xEdgeFace} in the following experiments.

\textbf{Model Complexity}: A key objective of this work is to focus on lightweight heterogeneous face recognition (HFR) models. Consequently, it is essential to compare the computational complexity of the proposed models against those commonly employed in existing literature. We evaluate the computational efficiency of our approach by reporting two key metrics: the number of floating-point operations (GFLOPs) and the total number of parameters (measured in millions, denoted as MPARAMs). As shown in Table~\ref{tab:computational_complexity}, the proposed xEdgeFace variants operate with significantly reduced computational overhead and parameter count, highlighting their suitability for deployment in resource-constrained environments.

\begin{table}[!htb]
  \caption{Comparison of computational complexity between the proposed method and state-of-the-art HFR approaches, reported in terms of floating point operations (GFLOPs) and number of parameters (MPARAMs). }
  \label{tab:computational_complexity}
  \centering
  \resizebox{0.7\columnwidth}{!}{%
\begin{tabular}{lcc}
\toprule
{} &  \textbf{GFLOPS} &  \textbf{MPARAMS} \\
\midrule
CAIM(1-3) \cite{george2023bridging} &    26.3 &    65.6 \\
DIU \cite{george2024heterogeneous} & 24.2 & 65.2 \\
SSMB \cite{george2024modality} &24.2 &65.5 \\
PDT \cite{george2022prepended} &24.2 &65.2 \\ \midrule
\rowcolor{Gray}
\textbf{xEdgeFace - Base} & 1.39 & 18.23 \\
\rowcolor{Gray}
\textbf{xEdgeFace - S} ($\gamma = 0.5$) & 0.31 & 3.65 \\
\rowcolor{Gray}
\textbf{xEdgeFace - XS} ($\gamma = 0.6$)  & 0.15 & 1.77 \\
\rowcolor{Gray}
\textbf{xEdgeFace - XXS} & 0.09 & 1.24 \\
\bottomrule
\end{tabular}
}
\end{table}

\textbf{Adapting Different Sets of Layers}: To determine the optimal set of layers for adaptation, we conduct a series of controlled ablation experiments. Specifically, we progressively adapt the LayerNorm (LN) layers, the initial convolutional stem (ST), and successive network stages: Stage 0 (S0), Stage 1 (S1), and Stage 2 (S2). The results, summarized in Table~\ref{tab:ablation_layer_tufts}, reveal that adapting only the LayerNorm layers yields a substantial performance gain over the baseline pretrained model. Further adaptation of the stem and early network stages continues to improve performance. However, the extent to which layers can be adapted is closely tied to the amount of available training data. Notably, configurations such as (LN, ST) and (LN, ST, S0) achieve a favorable trade-off, delivering strong performance while limiting the number of parameters adapted.

\begin{table}[h]
  \centering
  \caption{Ablation study on the Tufts Face Dataset using different configurations of adapted layers.}
  \label{tab:ablation_layer_tufts}
  \resizebox{0.47\textwidth}{!}{

\begin{tabular}{lcccc}
\toprule
\textbf{Adapted Layers} &    \textbf{AUC} &    \textbf{EER} &     \textbf{Rank-1} &   \textbf{VR@FAR=1$\%$} \\
\midrule
Baseline &  87.44  &  20.37 &  42.73 &  42.86 \\
LN &  96.68 &   9.09 &  73.97 &  77.18 \\
ST &  91.21 &  16.49 &  51.89 &   52.50 \\
LN, ST &  97.52 &   7.98 &  75.76 &  79.59 \\
LN, ST, S0 &  97.91 &   6.68 &  \textbf{82.59} &  \textbf{86.83} \\
LN, ST, S0, S1 &  \textbf{98.36} &   6.71 &  81.33 &   83.30 \\
LN, ST, S0, S1, S2 &  98.29 &   \textbf{6.64} &  79.53 &  84.23 \\
\bottomrule
\end{tabular}}
\end{table}

\textbf{Effect of Varying $\lambda$}: The hyperparameter $\lambda$ controls the trade-off between supervision from the pretrained model and modality alignment objective, both of which are critical for heterogeneous face recognition (HFR). The results of this ablation study are presented in Table~\ref{tab:ablation_lambda_tufts}. Setting $\lambda = 0$ emphasizes only the modality alignment objective, completely omitting supervision from the pretrained model. While this encourages domain adaptation, it leads to rapid overfitting due to the limited training data. Conversely, $\lambda = 1$ relies solely on the pretrained supervision, resulting in poor cross-modal performance due to insufficient domain alignment. Although both $\lambda$ = 0.50 and $\lambda$ = 0.75 perform very well, we find that $\lambda = 0.75$ offers the best balance, assigning greater importance to pretrained supervision given the relatively small size of the fine-tuning dataset. This configuration not only improves HFR performance but also mitigates catastrophic forgetting, preserving recognition performance in the original RGB domain (Table~\ref{tab:tufts_fr_perf}).

\begin{table}[h]
  \centering
  \caption{Ablation study with varying values of hyperparameter $\lambda$.}
  \label{tab:ablation_lambda_tufts}
  \resizebox{0.35\textwidth}{!}{
\begin{tabular}{lcccc}
\toprule
$\lambda$ &    \textbf{AUC} &    \textbf{EER} &     \textbf{Rank-1} & \textbf{VR@FAR=1$\%$} \\
\midrule
0.00 &  94.52 &  12.23 &  52.96 &  46.57 \\
0.25 &  97.87 &   7.24 &  79.53 &  81.82 \\
0.50 &  \textbf{98.46} &   \textbf{5.95} &   \textbf{83.3 }&  84.79 \\
0.75 &  97.91 &   6.68 &  82.59 &  \textbf{86.83} \\
1.00 &  87.47 &  20.22 &  42.19 &  42.86 \\
\bottomrule
\end{tabular}}
\end{table}

\textbf{Experiments with Other EdgeFace Variants}: While the main experiments were conducted using the EdgeFace-Base model, we also evaluated the effectiveness of our proposed adaptation on lighter variants of the architecture. Table~\ref{tab:arch_comp} presents a comparison between the original pretrained models and their adapted counterparts, denoted as xEdgeFace. The results indicate that the absolute heterogeneous face recognition (HFR) performance correlates with the performance of the original pretrained weights. However, our adaptation consistently yields significant improvements in all model sizes, achieving relative performance gains of 103\%, 194\%, 257\%, and 361\% from the largest to the most compact variant. These results highlight the scalability and generalizability of our method, demonstrating its ability to enhance HFR performance even in extremely lightweight architectures.

\begin{table}[h]
  \centering
  \caption{Comparison with different variants of EdgeFace}
  \label{tab:arch_comp}
  \resizebox{0.47\textwidth}{!}{
\begin{tabular}{lcccc}
\toprule
\textbf{Model} & \textbf{AUC} & \textbf{EER} & \textbf{Rank-1} & \textbf{VR@FAR=1\%} \\
\midrule
EdgeFace - Base & 87.44 & 20.37 & 42.73 & 42.86 \\
\rowcolor{Gray}
\textbf{xEdgeFace - Base} & 97.91 & 6.68 & 82.59 & \textbf{86.83} ($\uparrow\,\textbf{103}\%$) \\
EdgeFace - S ($\gamma = 0.5$) & 80.93 & 27.30 & 24.78 & 25.05 \\
\rowcolor{Gray}
\textbf{xEdgeFace - S} ($\gamma = 0.5$) & 96.93 & 8.89 & 71.10 & \textbf{73.65} ($\uparrow\,\textbf{194}\%$) \\
EdgeFace - XS ($\gamma = 0.6$) & 77.76 & 30.24 & 18.67 & 19.11 \\
\rowcolor{Gray}
\textbf{xEdgeFace - XS} ($\gamma = 0.6$) & 96.28 & 10.02 & 68.22 & \textbf{68.27} ($\uparrow\,\textbf{257}\%$) \\
EdgeFace - XXS & 75.72 & 31.73 & 17.41 & 12.80 \\
\rowcolor{Gray}
\textbf{xEdgeFace - XXS} & 95.14 & 11.35 & 60.50 & \textbf{59.00} ($\uparrow\,\textbf{361}\%$) \\
\bottomrule
\end{tabular}
}
\end{table}

\textbf{Face Recognition Performance of Adapted HFR Models}: To evaluate the face recognition (FR) performance of the adapted models beyond the HFR setting, we evaluate the xEdgeFace model on standard FR benchmarks. We evaluate the models on LFW~\cite{huang2008labeled}, CA-LFW~\cite{zheng2017cross}, CP-LFW~\cite{zheng2018cross}, CFP-FP~\cite{sengupta2016frontal}, and AgeDB-30~\cite{moschoglou2017agedb} and report the accuracies. We compare the xEdgeFace model adapted for both the VIS-NIR and VIS-Thermal HFR settings, with the latter presenting the most significant domain gap among all evaluated scenarios. As shown in Table~\ref{tab:tufts_fr_perf}, the adapted model achieves near-parity in face recognition performance with the original EdgeFace model, even under the challenging VIS-Thermal scenario. At the same time, xEdgeFace shows substantial gains in HFR performance, clearly demonstrating its capability to perform robustly across both homogeneous and heterogeneous FR tasks. This highlights the effectiveness of the self-distillation loss, which serves as a regularization mechanism that mitigates catastrophic forgetting and preserves the model’s discriminative ability in the original RGB domain. Essentially, the proposed training scheme extends the model's capabilities to heterogeneous face recognition without compromising its original performance, thereby enabling its effective use in both homogeneous and heterogeneous recognition tasks.

\begin{table}[h]
  \centering
  \caption{Face recognition performance of the pretrained and adapted model.}
  \label{tab:tufts_fr_perf}
  \resizebox{0.49\textwidth}{!}{
\begin{tabular}{lccccc}
\toprule

                  \textbf{Model} &           \textbf{LFW} ~\cite{huang2008labeled} &         \textbf{CALFW}~\cite{zheng2017cross} &         \textbf{CPLFW}~\cite{zheng2018cross} &        \textbf{CFP-FP}~\cite{sengupta2016frontal} &      \textbf{AGEDB-30}~\cite{moschoglou2017agedb} \\
\midrule

EdgeFace - Base	 & 99.83 ± 0.24	 & 96.07 ± 1.03	 & 93.75 ± 1.16	& 97.01 ± 0.94	& 97.60 ± 0.70 \\
\textbf{xEdgeFace - Base} (VIS-Thermal) &  99.78 ± 0.27 &  95.85 ± 1.16 &  93.62 ± 1.31 &  96.83 ± 0.99 &  97.50 ± 0.88  \\
\textbf{xEdgeFace - Base} (VIS-NIR) & 99.82 ± 0.26 & 96.07 ± 0.99 & 93.78 ± 1.24 & 96.94 ± 0.97 & 97.28 ± 0.83 \\
\bottomrule
\end{tabular}
}
\vspace{-2mm}
\end{table}

\subsection{Comparison with State-of-the-art}

In this section, we present a comparative evaluation of the proposed xEdgeFace model against state-of-the-art heterogeneous face recognition (HFR) methods reported in the literature. Notably, xEdgeFace is significantly more lightweight than the models it is compared against, highlighting the efficiency of our approach. For all experiments, we employ the xEdgeFace-Base variant with two adaptation configurations: (LN, ST) and (LN, ST, S0). The adaptation loss weight $\lambda$ is fixed at 0.75 across all evaluations to ensure consistency. However, this value can be further fine-tuned for individual datasets based on their size and face recognition performance requirements.

\textbf{Experiments with Tufts face dataset}: Table \ref{tab:tufts} presents the performance of xEdgeFace and other state-of-the-art methods on the VIS-Thermal protocol of the Tufts face dataset. This dataset poses significant challenges due to variations in pose and other factors, particularly extreme yaw angles, which degrade the performance of both visible-spectrum and heterogeneous face recognition systems. Despite these challenges, the xEdgeFace model with the (LN, ST, S0) configuration achieves the highest verification rate (69.02\%) and Rank-1 accuracy (82.59\%), despite being extremely lightweight.

\begin{table}[h]
  \centering
  \caption{Experimental results on VIS-Thermal protocol of the Tufts Face dataset.}
  \label{tab:tufts}
  \resizebox{0.99\columnwidth}{!}{
  \begin{tabular}{lccc}
    \toprule
    \textbf{Method} & \textbf{Rank-1} & \textbf{VR@FAR=1$\%$} & \textbf{VR@FAR=0.1$\%$}  \\ \midrule
      LightCNN \cite{Wu2018ALC} & 29.4 & 23.0 & 5.3 \\
      DVG \cite{fu2019dual} & 56.1 & 44.3 & 17.1 \\
      DVG-Face \cite{fu2021dvg} & 75.7& 68.5 & 36.5 \\ 
      DSU-Iresnet100 \cite{george2022prepended} & 49.7 & 49.8 & 28.3 \\   
      PDT \cite{george2022prepended} & 65.71 & 69.39 & 45.45 \\ 
      CAIM \cite{george2023bridging}  & 73.07 & 76.81 & 46.94 \\ 
      SSMB \cite{george2024modality} (N=1) & 75.04 & 78.29 & 53.99 \\
      SSMB \cite{george2024modality} (N=2) & 78.46 & 80.33   &54.55 \\
      \midrule
      \rowcolor{Gray}
      \textbf{xEdgeFace-Base (LN, ST)}  & 75.76 & 79.59 & 62.89 \\
      \rowcolor{Gray}
      \textbf{xEdgeFace-Base (LN, ST, S0)}  &\textbf{82.59} &\textbf{86.83} & \textbf{69.02} \\
      \bottomrule
  
  \end{tabular}
  }
\end{table}

\textbf{Experiments with MCXFace Dataset}: Table \ref{tab:mcxface} reports the average performance over five folds for the VIS-Thermal protocol on the MCXFace dataset. The results represent the mean and standard deviation values across all five folds. The baseline corresponds to the performance of the pretrained \textit{IresNet100} face recognition model evaluated directly on thermal images. As shown, the proposed xEdgeFace approach outperforms all compared methods, achieving the highest average Rank-1 accuracy of 91.68\%.

\begin{table}[h]
  \caption{Performance of the proposed approach in the VIS-Thermal protocol of  MCXFace dataset, the Baseline is a pretrained \textit{Iresnet100} model.}
  \label{tab:mcxface}
  \centering
  \resizebox{0.98\columnwidth}{!}{%
  \begin{tabular}{lrrr}
  \toprule
  \textbf{Method} & \textbf{AUC}   & \textbf{EER}   & \textbf{Rank-1}   \\ \midrule
 Baseline & 84.45 $\pm$ 3.70  & 22.07 $\pm$ 2.81 & 47.23 $\pm$ 3.93    \\
DSU-Iresnet100 \cite{george2022prepended} & 98.12 $\pm$ 0.75 & 6.58 $\pm$ 1.35 & 83.43 $\pm$ 5.47 \\
PDT \cite{george2022prepended}      & 98.43 $\pm$ 0.78  &  6.52 $\pm$ 1.45  & 84.52 $\pm$ 5.36   \\ 
CAIM \cite{george2023bridging}  & 98.97 $\pm$ 0.24& 5.05 $\pm$ 0.91& 87.24$\pm$2.75\\\midrule
\rowcolor{Gray}
\textbf{xEdgeFace-Base (LN, ST)} &  99.12$\pm$0.14 &  4.71$\pm$0.31 &  89.98$\pm$2.47 \\

\rowcolor{Gray}
\textbf{xEdgeFace-Base (LN, ST, S0)} &  99.50$\pm$0.21 &  3.42$\pm$0.78 &  \textbf{91.68$\pm$2.67}\\
\bottomrule
  \end{tabular}
  }
  \end{table}

\textbf{Experiments with Polathermal Dataset}: We performed experiments on the thermal-to-visible face recognition protocols of the Polathermal dataset, and the results are summarized in Table \ref{tab:polathermal}. The table reports the average Rank-1 identification accuracy across the five protocols defined in \cite{de2018heterogeneous}. Our proposed xEdgeFace approach achieves the highest performance, with an average Rank-1 accuracy of 97.31\%.

\begin{table}[ht]
\caption{Pola Thermal - Average Rank-1 recognition rate.}
\label{tab:polathermal}
\begin{center}
  \resizebox{0.7\columnwidth}{!}{
  \begin{tabular}{lr}
    \toprule
    \textbf{Method} & \textbf{Mean (Std. Dev.)} \\ \midrule
    
    DPM in \cite{hu2016polarimetric}   & 75.31 \% (-) \\ 
    CpNN in \cite{hu2016polarimetric}  & 78.72 \% (-) \\ 
    PLS in \cite{hu2016polarimetric}   & 53.05\% (-)  \\  \midrule

    LBPs + DoG in \cite{liao2009heterogeneous} & 36.8\% (3.5) \\ 
    ISV in \cite{de2016heterogeneous}       & 23.5\% (1.1) \\ 
    GFK in \cite{sequeira2017cross}             & 34.1\% (2.9) \\ 

    DSU(Best Result) \cite{de2018heterogeneous} & 76.3\% (2.1) \\

    DSU-Iresnet100 \cite{george2022prepended} & 88.2\% (5.8) \\
    PDT \cite{george2022prepended} & 97.1\% (1.3) \\ 
    
    CAIM \cite{george2023bridging} & 95.00\% (1.63) \\ \midrule
\rowcolor{Gray}
    \textbf{xEdgeFace-Base (LN, ST)} & 95.98\% (1.90) \\
    \rowcolor{Gray}
    \textbf{xEdgeFace-Base (LN, ST, S0)} & \textbf{97.31\% (1.96)} \\
    \bottomrule 
  \end{tabular}
  }
\end{center}
\end{table}

\vspace{-5mm}

\textbf{Experiments with SCFace Dataset}: The SCFace dataset presents a heterogeneity challenge due to the quality disparity between the gallery (high-resolution mugshots) and probe (low-resolution surveillance camera) images. Table \ref{tab:scface} presents the results, focusing on the ``far'' protocol, which is the most challenging among the evaluation settings. It can be seen that our approach achieved the best Rank-1 performance of 96.36\%. This indicates that aspects such as image quality degradation can also be addressed in our framework.

\begin{table}[h]
  \caption{Performance of the proposed approach in the SCFace dataset, performance reported on the \textit{far} protocol.}
  \label{tab:scface}
  \centering
  \resizebox{0.85\columnwidth}{!}{%
  \begin{tabular}{lrrrr}
  \toprule
               \textbf{Method} & \textbf{AUC}   & \textbf{EER}   & \textbf{Rank-1}    & \begin{tabular}[c]{@{}c@{}} \end{tabular} \\ \midrule

                                     DSU-Iresnet100 \cite{george2022prepended} & 97.18 & 8.37 & 79.53 \\

                              PDT \cite{george2022prepended}   &  98.31 & 6.98 &  84.19            \\ 
                               CAIM \cite{george2023bridging} &   98.81 &   5.09 &   86.05  \\

                             SSMB \cite{george2024modality} (N=1) & 98.77 &  5.91 &  87.73 \\
                             SSMB \cite{george2024modality} (N=2) & 98.67 & 6.36 & 86.82 \\ \midrule

                             \rowcolor{Gray}

                              \textbf{xEdgeFace-Base (LN, ST)}        &  \textbf{99.86} &  \textbf{1.82} &  \textbf{96.36} \\
                             \rowcolor{Gray}
                              \textbf{xEdgeFace-Base (LN, ST, S0)} &  99.71 &  2.27 &  93.18 \\
                
  \bottomrule
  \end{tabular}
  }
  \vspace{-2mm}

  \end{table}

\textbf{Experiments with CUFSF Dataset}: In this section, we evaluate our approach on the challenging task of sketch-to-photo face recognition. Table \ref{tab:cufsf} presents the Rank-1 accuracies achieved by the competing methods, following the protocols defined in \cite{fang2020identity}. Our method achieves a Rank-1 accuracy of 80.30\%, ranking second only to SSMB \cite{george2024modality}. However, overall performance in this modality remains lower compared to others, such as thermal or near-infrared, reflecting the inherent difficulty of the task. The CUFSF dataset consists of viewed hand-drawn sketches \cite{klum2014facesketchid}, which, while appearing visually similar to the original subjects to humans, often lack the discriminative features critical for face recognition models. Unlike other imaging modalities, sketches are influenced by artistic interpretation and exaggeration, further widening the domain gap. Nonetheless, our method demonstrates promising performance, highlighting its robustness even under such extreme modality shifts.

\begin{table}[ht]
\caption{CUFSF: Rank-1 recognition rate in sketch to photo recognition.}
\label{tab:cufsf}

\begin{center}
\resizebox{0.49\columnwidth}{!}{%
  \begin{tabular}{lrr}
    \toprule
    \textbf{Method} & \textbf{Rank-1} \\ \midrule
    IACycleGAN \cite{fang2020identity} &64.94 \\
    DSU-Iresnet100 \cite{george2022prepended} & 67.06 \\ 
    PDT \cite{george2022prepended}  & 71.08 \\ 
    CAIM \cite{george2023bridging} & 76.38 \\ 
    SSMB \cite{george2024modality} (N=1) &  81.14\\
    SSMB \cite{george2024modality} (N=2) & \textbf{81.67} \\ \midrule
\rowcolor{Gray}
    \textbf{xEdgeFace-Base (LN, ST)} & 80.30 \\
\rowcolor{Gray}
    \textbf{xEdgeFace-Base (LN, ST, S0)} & 78.81 \\
    \bottomrule 
  \end{tabular}
  }
\end{center}
\vspace{-3mm}
\end{table}

\textbf{Experiments with CASIA-VIS-NIR 2.0 Dataset}: We evaluate the proposed method on the CASIA NIR-VIS 2.0 dataset to assess its VIS-NIR matching performance. Owing to the relatively small domain gap, VIS-pretrained models already achieve competitive performance. To enable rigorous comparison, we adopt stricter evaluation metrics: VR@FAR=0.1\% and 0.01\%. Following the standard protocol, we report average performance and standard deviation across 10 folds. As shown in Table \ref{tab:casia}, our method consistently outperforms existing state-of-the-art approaches, demonstrating strong generalization capabilities.
\begin{table}[h]
  \centering
  \caption{Experimental results on CASIA NIR-VIS 2.0.}
  \label{tab:casia}
  \resizebox{0.47\textwidth}{!}{
  \begin{tabular}{l|ccc}
    \toprule
      \textbf{Method} & \textbf{Rank-1} & \textbf{VR@FAR=0.1$\%$} & \textbf{VR@FAR=0.01$\%$} \\
      \midrule
      IDNet \cite{Reale2016SeeingTF} & 87.1$\pm$0.9 & 74.5 & - \\
      HFR-CNN \cite{saxena2016heterogeneous} & 85.9$\pm$0.9 & 78.0 & - \\
      Hallucination \cite{Lezama2017NotAO} & 89.6$\pm$0.9 & - & - \\
      TRIVET \cite{XXLiu:2016} & 95.7$\pm$0.5 & 91.0$\pm$1.3 & 74.5$\pm$0.7 \\
      W-CNN \cite{DBLP:journals/corr/abs-1708-02412} & 98.7$\pm$0.3 & 98.4$\pm$0.4 & 94.3$\pm$0.4 \\
      PACH \cite{duan2019pose} & 98.9$\pm$0.2 & 98.3$\pm$0.2 & - \\
      RCN \cite{deng2019residual} & 99.3$\pm$0.2 & 98.7$\pm$0.2 & - \\
      MC-CNN \cite{8624555} & 99.4$\pm$0.1 & 99.3$\pm$0.1 & - \\
      DVR \cite{XWu:2019}  & 99.7$\pm$0.1 & 99.6$\pm$0.3 & 98.6$\pm$0.3 \\
      DVG \cite{fu2019dual} & 99.8$\pm$0.1 & 99.8$\pm$0.1 & 98.8$\pm$0.2 \\
      DVG-Face \cite{fu2021dvg} & 99.9$\pm$0.1 & 99.9$\pm$0.0 & 99.2$\pm$0.1 \\     
      PDT \cite{george2022prepended} & 99.95$\pm$0.04 & 99.94$\pm$0.03 & 99.77$\pm$0.09 \\ 
      CAIM \cite{george2023bridging} & 99.96$\pm$0.02 &\textbf{99.95$\pm$0.02} & 99.79$\pm$0.11 \\ \midrule
      \rowcolor{Gray}
      \textbf{xEdgeFace-Base (LN, ST)} & 99.96$\pm$0.02 & 99.91$\pm$0.02 & 99.83$\pm$0.04 \\
      \rowcolor{Gray}
      \textbf{xEdgeFace-Base (LN, ST, S0)} & \textbf{99.99$\pm$0.01} & 99.93$\pm$0.02 & \textbf{99.86$\pm$0.04}\\
      \bottomrule
  \end{tabular}}
\end{table}

\vspace{-3mm}

\textbf{Discussions}: The extensive experimental results in six different heterogeneous face recognition (HFR) benchmarks demonstrate the effectiveness, generalizability, and efficiency of the proposed xEdgeFace framework. Despite its lightweight nature, xEdgeFace consistently outperforms or closely matches state-of-the-art methods across modalities, including thermal, NIR, sketch, and low-resolution surveillance images. In particular, our approach achieves top Rank-1 accuracy in challenging datasets while maintaining minimal degradation in standard face recognition benchmarks. Ablation studies show insights into which components are most effective for adaptation and the effectiveness of self-distillation in balancing domain alignment and the retention of pretrained knowledge. The framework also scales effectively to extremely compact model variants, achieving significant relative gains, which highlights its suitability for edge deployment. Furthermore, strong performance under large domain gaps (e.g., sketch-photo) confirms the robustness of our adaptation strategy across challenging heterogeneous settings.
\section{Conclusions}
In this work, we proposed xEdgeFace, an efficient framework for lightweight HFR that extends existing FR models for cross-modal scenarios. By selectively adapting early convolutional and layer normalization (LN) layers within a contrastive self-distillation framework, our method enables strong cross-modal generalization while preserving the model’s original capabilities in the visible spectrum. This design ensures that the adapted model performs robustly on diverse and challenging HFR tasks such as VIS-Thermal, VIS-NIR, and sketch-photo recognition, but also maintains competitive performance on standard FR benchmarks, effectively mitigating catastrophic forgetting. Extensive experiments show that xEdgeFace consistently outperforms or matches state-of-the-art methods, even with highly compact model variants, making it well-suited for edge deployment. We will publicly release our code base and protocols, enabling further extension of our work.
\section{Acknowledgements}
This research was funded by the European Union project CarMen (Grant Agreement No. 101168325) as well as the Swiss Center for Biometrics Research and Testing.

{\small
\bibliographystyle{ieee}
\bibliography{egbib}

\begin{thebibliography}{10}\itemsep=-1pt

\bibitem{alansari2023ghostfacenets}
M.~Alansari, O.~A. Hay, S.~Javed, A.~Shoufan, Y.~Zweiri, and N.~Werghi.
\newblock Ghostfacenets: Lightweight face recognition model from cheap
  operations.
\newblock {\em IEEE Access}, 2023.

\bibitem{anghelone2025beyond}
D.~Anghelone, C.~Chen, A.~Ross, and A.~Dantcheva.
\newblock Beyond the visible: A survey on cross-spectral face recognition.
\newblock {\em Neurocomputing}, 611:128626, 2025.

\bibitem{bob2017}
A.~Anjos, M.~G\"unther, T.~de~Freitas~Pereira, P.~Korshunov, A.~Mohammadi, and
  S.~Marcel.
\newblock Continuously reproducing toolchains in pattern recognition and
  machine learning experiments.
\newblock In {\em International Conference on Machine Learning (ICML)}, Aug.
  2017.

\bibitem{bob2012}
A.~Anjos, L.~E. Shafey, R.~Wallace, M.~G\"unther, C.~McCool, and S.~Marcel.
\newblock Bob: a free signal processing and machine learning toolbox for
  researchers.
\newblock In {\em 20th ACM Conference on Multimedia Systems (ACMMM), Nara,
  Japan}, Oct. 2012.

\bibitem{ba2016layer}
J.~L. Ba, J.~R. Kiros, and G.~E. Hinton.
\newblock Layer normalization.
\newblock {\em arXiv preprint arXiv:1607.06450}, 2016.

\bibitem{boutros2021mixfacenets}
F.~Boutros, N.~Damer, M.~Fang, F.~Kirchbuchner, and A.~Kuijper.
\newblock Mixfacenets: Extremely efficient face recognition networks.
\newblock In {\em 2021 IEEE International Joint Conference on Biometrics
  (IJCB)}, pages 1--8. IEEE, 2021.

\bibitem{boutros2022pocketnet}
F.~Boutros, P.~Siebke, M.~Klemt, N.~Damer, F.~Kirchbuchner, and A.~Kuijper.
\newblock Pocketnet: Extreme lightweight face recognition network using neural
  architecture search and multistep knowledge distillation.
\newblock {\em IEEE Access}, 10:46823--46833, 2022.

\bibitem{chen2018mobilefacenets}
S.~Chen, Y.~Liu, X.~Gao, and Z.~Han.
\newblock Mobilefacenets: Efficient cnns for accurate real-time face
  verification on mobile devices.
\newblock In {\em Biometric Recognition: 13th Chinese Conference, CCBR 2018,
  Urumqi, China, August 11-12, 2018, Proceedings 13}, pages 428--438. Springer,
  2018.

\bibitem{de2018heterogeneous}
T.~de~Freitas~Pereira, A.~Anjos, and S.~Marcel.
\newblock Heterogeneous face recognition using domain specific units.
\newblock {\em IEEE Transactions on Information Forensics and Security},
  14(7):1803--1816, 2018.

\bibitem{de2016heterogeneous}
T.~de~Freitas~Pereira and S.~Marcel.
\newblock Heterogeneous face recognition using inter-session variability
  modelling.
\newblock In {\em Proceedings of the IEEE Conference on Computer Vision and
  Pattern Recognition Workshops}, pages 111--118, 2016.

\bibitem{deng2019lightweight}
J.~Deng, J.~Guo, D.~Zhang, Y.~Deng, X.~Lu, and S.~Shi.
\newblock Lightweight face recognition challenge.
\newblock In {\em Proceedings of the IEEE/CVF International Conference on
  Computer Vision Workshops}, pages 0--0, 2019.

\bibitem{8624555}
Z.~Deng, X.~Peng, Z.~Li, and Y.~Qiao.
\newblock Mutual component convolutional neural networks for heterogeneous face
  recognition.
\newblock {\em IEEE Transactions on Image Processing}, 28(6):3102--3114, 2019.

\bibitem{deng2019residual}
Z.~Deng, X.~Peng, and Y.~Qiao.
\newblock Residual compensation networks for heterogeneous face recognition.
\newblock In {\em AAAI Conference on Artificial Intelligence}, 2019.

\bibitem{duan2019pose}
B.~Duan, C.~Fu, Y.~Li, X.~Song, and R.~He.
\newblock Pose agnostic cross-spectral hallucination via disentangling
  independent factors.
\newblock In {\em IEEE Conference on Computer Vision and Pattern Recognition},
  2020.

\bibitem{fang2020identity}
Y.~Fang, W.~Deng, J.~Du, and J.~Hu.
\newblock {Identity-aware {CycleGAN} for face photo-sketch synthesis and
  recognition}.
\newblock {\em Pattern Recognition}, 102:107249, 2020.

\bibitem{fu2019dual}
C.~Fu, X.~Wu, Y.~Hu, H.~Huang, and R.~He.
\newblock Dual variational generation for low shot heterogeneous face
  recognition.
\newblock In {\em Advances in Neural Information Processing Systems}, 2019.

\bibitem{fu2021dvg}
C.~Fu, X.~Wu, Y.~Hu, H.~Huang, and R.~He.
\newblock {DVG-face}: Dual variational generation for heterogeneous face
  recognition.
\newblock {\em IEEE Transactions on Pattern Analysis and Machine Intelligence},
  2021.

\bibitem{gatys2016image}
L.~A. Gatys, A.~S. Ecker, and M.~Bethge.
\newblock Image style transfer using convolutional neural networks.
\newblock In {\em Proceedings of the IEEE conference on computer vision and
  pattern recognition}, pages 2414--2423, 2016.

\bibitem{george2023edgeface}
A.~George, C.~Ecabert, H.~O. Shahreza, K.~Kotwal, and S.~Marcel.
\newblock {EdgeFace}: Efficient face recognition model for edge devices.
\newblock {\em arXiv preprint arXiv:2307.01838}, 2023.

\bibitem{george2024edgeface}
A.~George, C.~Ecabert, H.~O. Shahreza, K.~Kotwal, and S.~Marcel.
\newblock Edgeface: Efficient face recognition model for edge devices.
\newblock {\em IEEE Transactions on Biometrics, Behavior, and Identity
  Science}, 6(2):158--168, 2024.

\bibitem{george2022comprehensive}
A.~George, D.~Geissbuhler, and S.~Marcel.
\newblock A comprehensive evaluation on multi-channel biometric face
  presentation attack detection.
\newblock {\em arXiv preprint arXiv:2202.10286}, 2022.

\bibitem{george2023bridging}
A.~George and S.~Marcel.
\newblock Bridging the gap: Heterogeneous face recognition with conditional
  adaptive instance modulation.
\newblock In {\em 2023 IEEE International Joint Conference on Biometrics
  (IJCB)}, pages 1--10. IEEE, 2023.

\bibitem{george2024modalities}
A.~George and S.~Marcel.
\newblock From modalities to styles: Rethinking the domain gap in heterogeneous
  face recognition.
\newblock {\em IEEE Transactions on Biometrics, Behavior, and Identity
  Science}, 6(4):475--485, 2024.

\bibitem{george2024heterogeneous}
A.~George and S.~Marcel.
\newblock Heterogeneous face recognition using domain invariant units.
\newblock In {\em ICASSP 2024-2024 IEEE International Conference on Acoustics,
  Speech and Signal Processing (ICASSP)}, pages 4780--4784. IEEE, 2024.

\bibitem{george2024modality}
A.~George and S.~Marcel.
\newblock Modality agnostic heterogeneous face recognition with switch style
  modulators.
\newblock In {\em 2024 IEEE International Joint Conference on Biometrics
  (IJCB)}, pages 1--10. IEEE, 2024.

\bibitem{george2025digi2real}
A.~George and S.~Marcel.
\newblock Digi2real: Bridging the realism gap in synthetic data face
  recognition via foundation models.
\newblock In {\em Proceedings of the Winter Conference on Applications of
  Computer Vision}, pages 1469--1478, 2025.

\bibitem{george2022prepended}
A.~George, A.~Mohammadi, and S.~Marcel.
\newblock Prepended domain transformer: Heterogeneous face recognition without
  bells and whistles.
\newblock {\em IEEE Transactions on Information Forensics and Security}, 2022.

\bibitem{grgic2011scface}
M.~Grgic, K.~Delac, and S.~Grgic.
\newblock {SCface}--surveillance cameras face database.
\newblock {\em Multimedia tools and applications}, 51(3):863--879, 2011.

\bibitem{han2020model}
K.~Han, Y.~Wang, Q.~Zhang, W.~Zhang, C.~Xu, and T.~Zhang.
\newblock Model rubik’s cube: Twisting resolution, depth and width for
  tinynets.
\newblock {\em Advances in Neural Information Processing Systems},
  33:19353--19364, 2020.

\bibitem{happy2012video}
S.~Happy, A.~Dasgupta, A.~George, and A.~Routray.
\newblock A video database of human faces under near infra-red illumination for
  human computer interaction applications.
\newblock In {\em 2012 4th International Conference on Intelligent Human
  Computer Interaction (IHCI)}, pages 1--4. IEEE, 2012.

\bibitem{he2017learning}
R.~He, X.~Wu, Z.~Sun, and T.~Tan.
\newblock Learning invariant deep representation for {Nir-Vis} face
  recognition.
\newblock In {\em Thirty-First AAAI Conference on Artificial Intelligence},
  2017.

\bibitem{he2018wasserstein}
R.~He, X.~Wu, Z.~Sun, and T.~Tan.
\newblock Wasserstein {CNN}: Learning invariant features for {Nir-Vis} face
  recognition.
\newblock {\em IEEE transactions on pattern analysis and machine intelligence},
  41(7):1761--1773, 2018.

\bibitem{DBLP:journals/corr/abs-1708-02412}
R.~He, X.~Wu, Z.~Sun, and T.~Tan.
\newblock Wasserstein {CNN}: Learning invariant features for {Nir-Vis} face
  recognition.
\newblock {\em IEEE Transactions on Pattern Analysis and Machine Intelligence},
  41(7):1761--1773, 2018.

\bibitem{howard2017mobilenets}
A.~G. Howard, M.~Zhu, B.~Chen, D.~Kalenichenko, W.~Wang, T.~Weyand,
  M.~Andreetto, and H.~Adam.
\newblock Mobilenets: Efficient convolutional neural networks for mobile vision
  applications.
\newblock {\em arXiv preprint arXiv:1704.04861}, 2017.

\bibitem{hu2022lora}
E.~J. Hu, Y.~Shen, P.~Wallis, Z.~Allen-Zhu, Y.~Li, S.~Wang, L.~Wang, W.~Chen,
  et~al.
\newblock Lora: Low-rank adaptation of large language models.
\newblock {\em ICLR}, 1(2):3, 2022.

\bibitem{hu2016polarimetric}
S.~Hu, N.~J. Short, B.~S. Riggan, C.~Gordon, K.~P. Gurton, M.~Thielke,
  P.~Gurram, and A.~L. Chan.
\newblock A polarimetric thermal database for face recognition research.
\newblock In {\em Proceedings of the IEEE conference on computer vision and
  pattern recognition workshops}, pages 119--126, 2016.

\bibitem{huang2008labeled}
G.~B. Huang, M.~Mattar, T.~Berg, and E.~Learned-Miller.
\newblock Labeled faces in the wild: A database forstudying face recognition in
  unconstrained environments.
\newblock In {\em Workshop on faces in'Real-Life'Images: detection, alignment,
  and recognition}, 2008.

\bibitem{khan2022transformers}
S.~Khan, M.~Naseer, M.~Hayat, S.~W. Zamir, F.~S. Khan, and M.~Shah.
\newblock Transformers in vision: A survey.
\newblock {\em ACM computing surveys (CSUR)}, 54(10s):1--41, 2022.

\bibitem{klare2010matching}
B.~Klare, Z.~Li, and A.~K. Jain.
\newblock Matching forensic sketches to mug shot photos.
\newblock {\em IEEE transactions on pattern analysis and machine intelligence},
  33(3):639--646, 2010.

\bibitem{klare2012heterogeneous}
B.~F. Klare and A.~K. Jain.
\newblock Heterogeneous face recognition using kernel prototype similarities.
\newblock {\em IEEE transactions on pattern analysis and machine intelligence},
  35(6):1410--1422, 2012.

\bibitem{klum2014facesketchid}
S.~J. Klum, H.~Han, B.~F. Klare, and A.~K. Jain.
\newblock The facesketchid system: Matching facial composites to mugshots.
\newblock {\em IEEE Transactions on Information Forensics and Security},
  9(12):2248--2263, 2014.

\bibitem{kolf2023efar}
J.~N. Kolf, F.~Boutros, J.~Elliesen, M.~Theuerkauf, N.~Damer, M.~Alansari,
  O.~A. Hay, S.~Alansari, S.~Javed, N.~Werghi, et~al.
\newblock Efar 2023: Efficient face recognition competition.
\newblock In {\em 2023 IEEE International Joint Conference on Biometrics
  (IJCB)}, pages 1--12. IEEE, 2023.

\bibitem{learned2016labeled}
E.~Learned-Miller, G.~B. Huang, A.~RoyChowdhury, H.~Li, and G.~Hua.
\newblock Labeled faces in the wild: A survey.
\newblock {\em Advances in face detection and facial image analysis},
  1:189--248, 2016.

\bibitem{lei2009coupled}
Z.~Lei and S.~Z. Li.
\newblock Coupled spectral regression for matching heterogeneous faces.
\newblock In {\em 2009 IEEE Conference on Computer Vision and Pattern
  Recognition}, pages 1123--1128. IEEE, 2009.

\bibitem{Lezama2017NotAO}
J.~Lezama, Q.~Qiu, and G.~Sapiro.
\newblock Not afraid of the dark: {Nir-Vis} face recognition via cross-spectral
  hallucination and low-rank embedding.
\newblock In {\em IEEE Conference on Computer Vision and Pattern Recognition},
  2017.

\bibitem{li2013casia}
S.~Li, D.~Yi, Z.~Lei, and S.~Liao.
\newblock The {CASIA Nir-Vis 2.0} face database.
\newblock In {\em Proceedings of the IEEE conference on computer vision and
  pattern recognition workshops}, pages 348--353, 2013.

\bibitem{li2007illumination}
S.~Z. Li, R.~Chu, S.~Liao, and L.~Zhang.
\newblock Illumination invariant face recognition using near-infrared images.
\newblock {\em IEEE Transactions on pattern analysis and machine intelligence},
  29(4):627--639, 2007.

\bibitem{liao2009heterogeneous}
S.~Liao, D.~Yi, Z.~Lei, R.~Qin, and S.~Z. Li.
\newblock Heterogeneous face recognition from local structures of normalized
  appearance.
\newblock In {\em International Conference on Biometrics}, pages 209--218.
  Springer, 2009.

\bibitem{liu2021heterogeneous}
D.~Liu, X.~Gao, C.~Peng, N.~Wang, and J.~Li.
\newblock Heterogeneous face interpretable disentangled representation for
  joint face recognition and synthesis.
\newblock {\em IEEE transactions on neural networks and learning systems},
  33(10):5611--5625, 2021.

\bibitem{liu2020coupled}
D.~Liu, X.~Gao, N.~Wang, J.~Li, and C.~Peng.
\newblock Coupled attribute learning for heterogeneous face recognition.
\newblock {\em IEEE Transactions on Neural Networks and Learning Systems},
  31(11):4699--4712, 2020.

\bibitem{liu2018composite}
D.~Liu, J.~Li, N.~Wang, C.~Peng, and X.~Gao.
\newblock Composite components-based face sketch recognition.
\newblock {\em Neurocomputing}, 302:46--54, 2018.

\bibitem{liu2023modality}
D.~Liu, W.~Yang, C.~Peng, N.~Wang, R.~Hu, and X.~Gao.
\newblock Modality-agnostic augmented multi-collaboration representation for
  semi-supervised heterogenous face recognition.
\newblock In {\em Proceedings of the 31st ACM International Conference on
  Multimedia}, pages 4647--4656, 2023.

\bibitem{liu2005nonlinear}
Q.~Liu, X.~Tang, H.~Jin, H.~Lu, and S.~Ma.
\newblock A nonlinear approach for face sketch synthesis and recognition.
\newblock In {\em 2005 IEEE Computer Society conference on computer vision and
  pattern recognition (CVPR'05)}, volume~1, pages 1005--1010. IEEE, 2005.

\bibitem{XXLiu:2016}
X.~Liu, L.~Song, X.~Wu, and T.~Tan.
\newblock Transferring deep representation for {Nir-Vis} heterogeneous face
  recognition.
\newblock In {\em International Conference on Biometrics}, 2016.

\bibitem{luo2022memory}
M.~Luo, H.~Wu, H.~Huang, W.~He, and R.~He.
\newblock Memory-modulated transformer network for heterogeneous face
  recognition.
\newblock {\em IEEE Transactions on Information Forensics and Security},
  17:2095--2109, 2022.

\bibitem{ma2018shufflenet}
N.~Ma, X.~Zhang, H.-T. Zheng, and J.~Sun.
\newblock Shufflenet v2: Practical guidelines for efficient cnn architecture
  design.
\newblock In {\em Proceedings of the European conference on computer vision
  (ECCV)}, pages 116--131, 2018.

\bibitem{maaz2023edgenext}
M.~Maaz, A.~Shaker, H.~Cholakkal, S.~Khan, S.~W. Zamir, R.~M. Anwer, and
  F.~Shahbaz~Khan.
\newblock Edgenext: efficiently amalgamated cnn-transformer architecture for
  mobile vision applications.
\newblock In {\em Computer Vision--ECCV 2022 Workshops: Tel Aviv, Israel,
  October 23--27, 2022, Proceedings, Part VII}, pages 3--20. Springer, 2023.

\bibitem{martindez2019shufflefacenet}
Y.~Martindez-Diaz, L.~S. Luevano, H.~Mendez-Vazquez, M.~Nicolas-Diaz, L.~Chang,
  and M.~Gonzalez-Mendoza.
\newblock Shufflefacenet: A lightweight face architecture for efficient and
  highly-accurate face recognition.
\newblock In {\em Proceedings of the IEEE/CVF International Conference on
  Computer Vision Workshops}, pages 0--0, 2019.

\bibitem{moschoglou2017agedb}
S.~Moschoglou, A.~Papaioannou, C.~Sagonas, J.~Deng, I.~Kotsia, and
  S.~Zafeiriou.
\newblock Agedb: the first manually collected, in-the-wild age database.
\newblock In {\em proceedings of the IEEE conference on computer vision and
  pattern recognition workshops}, pages 51--59, 2017.

\bibitem{mostaani2020high}
Z.~Mostaani, A.~George, G.~Heusch, D.~Geissbuhler, and S.~Marcel.
\newblock The high-quality wide multi-channel attack (hq-wmca) database.
\newblock {\em arXiv preprint arXiv:2009.09703}, 2020.

\bibitem{panetta2018comprehensive}
K.~Panetta, Q.~Wan, S.~Agaian, S.~Rajeev, S.~Kamath, R.~Rajendran, S.~P. Rao,
  A.~Kaszowska, H.~A. Taylor, A.~Samani, et~al.
\newblock A comprehensive database for benchmarking imaging systems.
\newblock {\em IEEE transactions on pattern analysis and machine intelligence},
  42(3):509--520, 2018.

\bibitem{phillips1998feret}
P.~J. Phillips, H.~Wechsler, J.~Huang, and P.~J. Rauss.
\newblock The {FERET} database and evaluation procedure for face-recognition
  algorithms.
\newblock {\em Image and vision computing}, 16(5):295--306, 1998.

\bibitem{Reale2016SeeingTF}
C.~Reale, N.~M. Nasrabadi, H.~Kwon, and R.~Chellappa.
\newblock Seeing the forest from the trees: A holistic approach to
  near-infrared heterogeneous face recognition.
\newblock In {\em IEEE Conference on Computer Vision and Pattern Recognition
  Workshops}, 2016.

\bibitem{roy2018novel}
H.~Roy and D.~Bhattacharjee.
\newblock A novel quaternary pattern of local maximum quotient for
  heterogeneous face recognition.
\newblock {\em Pattern Recognition Letters}, 113:19--28, 2018.

\bibitem{sandler2018mobilenetv2}
M.~Sandler, A.~Howard, M.~Zhu, A.~Zhmoginov, and L.-C. Chen.
\newblock Mobilenetv2: Inverted residuals and linear bottlenecks.
\newblock In {\em Proceedings of the IEEE conference on computer vision and
  pattern recognition}, pages 4510--4520, 2018.

\bibitem{saxena2016heterogeneous}
S.~Saxena and J.~Verbeek.
\newblock Heterogeneous face recognition with {CNNs}.
\newblock In {\em European Conference on Computer Vision}, 2016.

\bibitem{sengupta2016frontal}
S.~Sengupta, J.-C. Chen, C.~Castillo, V.~M. Patel, R.~Chellappa, and D.~W.
  Jacobs.
\newblock Frontal to profile face verification in the wild.
\newblock In {\em 2016 IEEE winter conference on applications of computer
  vision (WACV)}, pages 1--9. IEEE, 2016.

\bibitem{sequeira2017cross}
A.~F. Sequeira, L.~Chen, J.~Ferryman, P.~Wild, F.~Alonso-Fernandez, J.~Bigun,
  K.~B. Raja, R.~Raghavendra, C.~Busch, T.~de~Freitas~Pereira, et~al.
\newblock Cross-eyed 2017: Cross-spectral iris/periocular recognition
  competition.
\newblock In {\em 2017 IEEE International Joint Conference on Biometrics
  (IJCB)}, pages 725--732. IEEE, 2017.

\bibitem{shahreza2024knowledge}
H.~O. Shahreza, A.~George, and S.~Marcel.
\newblock Knowledge distillation for face recognition using synthetic data with
  dynamic latent sampling.
\newblock {\em IEEE Access}, 2024.

\bibitem{sharma2011bypassing}
A.~Sharma and D.~W. Jacobs.
\newblock Bypassing synthesis: {PLS} for face recognition with pose,
  low-resolution and sketch.
\newblock In {\em CVPR 2011}, pages 593--600. IEEE, 2011.

\bibitem{tan2019mixconv}
M.~Tan and Q.~V. Le.
\newblock Mixconv: Mixed depthwise convolutional kernels.
\newblock {\em arXiv preprint arXiv:1907.09595}, 2019.

\bibitem{wang2008face}
X.~Wang and X.~Tang.
\newblock Face photo-sketch synthesis and recognition.
\newblock {\em IEEE transactions on pattern analysis and machine intelligence},
  31(11):1955--1967, 2008.

\bibitem{wu2018shift}
B.~Wu, A.~Wan, X.~Yue, P.~Jin, S.~Zhao, N.~Golmant, A.~Gholaminejad,
  J.~Gonzalez, and K.~Keutzer.
\newblock Shift: A zero flop, zero parameter alternative to spatial
  convolutions.
\newblock In {\em Proceedings of the IEEE conference on computer vision and
  pattern recognition}, pages 9127--9135, 2018.

\bibitem{Wu2018ALC}
X.~Wu, R.~He, Z.~Sun, and T.~Tan.
\newblock A light {CNN} for deep face representation with noisy labels.
\newblock {\em IEEE Transactions on Information Forensics and Security},
  13(11):2884--2896, 2018.

\bibitem{XWu:2019}
X.~Wu, H.~Huang, V.~M. Patel, R.~He, and Z.~Sun.
\newblock Disentangled variational representation for heterogeneous face
  recognition.
\newblock In {\em AAAI Conference on Artificial Intelligence}, 2019.

\bibitem{yan2019vargfacenet}
M.~Yan, M.~Zhao, Z.~Xu, Q.~Zhang, G.~Wang, and Z.~Su.
\newblock Vargfacenet: An efficient variable group convolutional neural network
  for lightweight face recognition.
\newblock In {\em Proceedings of the IEEE/CVF International Conference on
  Computer Vision Workshops}, pages 0--0, 2019.

\bibitem{yi2014learning}
D.~Yi, Z.~Lei, S.~Liao, and S.~Z. Li.
\newblock Learning face representation from scratch.
\newblock {\em arXiv preprint arXiv:1411.7923}, 2014.

\bibitem{yi2007face}
D.~Yi, R.~Liu, R.~Chu, Z.~Lei, and S.~Z. Li.
\newblock Face matching between near infrared and visible light images.
\newblock In {\em International Conference on Biometrics}, pages 523--530.
  Springer, 2007.

\bibitem{zhang2017generative}
H.~Zhang, V.~M. Patel, B.~S. Riggan, and S.~Hu.
\newblock Generative adversarial network-based synthesis of visible faces from
  polarimetric thermal faces.
\newblock In {\em 2017 IEEE International Joint Conference on Biometrics
  (IJCB)}, pages 100--107. IEEE, 2017.

\bibitem{zhang2011coupled}
W.~Zhang, X.~Wang, and X.~Tang.
\newblock Coupled information-theoretic encoding for face photo-sketch
  recognition.
\newblock In {\em CVPR 2011}, pages 513--520. IEEE, 2011.

\bibitem{zhao2024tuning}
B.~Zhao, H.~Tu, C.~Wei, J.~Mei, and C.~Xie.
\newblock Tuning layernorm in attention: Towards efficient multi-modal {LLM}
  finetuning.
\newblock In {\em The Twelfth International Conference on Learning
  Representations}, 2024.

\bibitem{zheng2018cross}
T.~Zheng and W.~Deng.
\newblock Cross-pose lfw: A database for studying cross-pose face recognition
  in unconstrained environments.
\newblock {\em Beijing University of Posts and Telecommunications, Tech. Rep},
  5(7), 2018.

\bibitem{zheng2017cross}
T.~Zheng, W.~Deng, and J.~Hu.
\newblock Cross-age lfw: A database for studying cross-age face recognition in
  unconstrained environments.
\newblock {\em arXiv preprint arXiv:1708.08197}, 2017.

\bibitem{zhuUnpairedImagetoImageTranslation2017}
J.-Y. Zhu, T.~Park, P.~Isola, and A.~A. Efros.
\newblock Unpaired {{Image-to-Image Translation}} using {{Cycle-Consistent
  Adversarial Networks}}.
\newblock {\em arXiv:1703.10593 [cs]}, Mar. 2017.

\bibitem{zhu2021webface260m}
Z.~Zhu, G.~Huang, J.~Deng, Y.~Ye, J.~Huang, X.~Chen, J.~Zhu, T.~Yang, J.~Lu,
  D.~Du, et~al.
\newblock Webface260m: A benchmark unveiling the power of million-scale deep
  face recognition.
\newblock In {\em Proceedings of the IEEE/CVF Conference on Computer Vision and
  Pattern Recognition}, pages 10492--10502, 2021.

\end{thebibliography}
}

\end{document}